\documentclass[twocolumn]{article}

\usepackage{epsfig}
\usepackage{graphicx}
\usepackage{amsmath}
\usepackage{amssymb}
\usepackage{subfig}
\usepackage{url}

\def\bw{{\mathbf w}}

\def\bx{{\mathbf x}}
\def\by{{\mathbf y}}

\def\balpha{{\mathbf \alpha}}
\def\blambda{{\mathbf \lambda}}

\def\bone{{\mathbf 1}}
\def\bzero{{\mathbf 0}}
\def\bK{{\mathbf K}}
\def\bX{{\mathbf X}}
\def\bY{{\mathbf Y}}
\def\bQ{{\mathbf Q}}
\def\bA{{\mathbf A}}
\def\RR{{\mathbb R}}
\def\bgamma{{\mathbf \gamma}}

\begin{document}
\date{}
\title{\large \bf Thesis: Multiple Kernel Learning for Object Categorization}
\author{Dinesh Govindaraj \\ Advisor: Prof. Chiranjib Bhattacharyya \\Submission:\ 2009 \\ Indian Institute of Science, Bangalore.}
\maketitle
\thispagestyle{empty}
\bibliographystyle{unsrt}
\begin{abstract}
Object Categorization is a challenging problem, especially when the images have clutter background, occlusions or different lighting conditions. In the past, many descriptors have been proposed which aid object categorization even in such adverse conditions. Each descriptor has its own merits and de-merits. Some descriptors are invariant to transformations while the others are more discriminative~\cite{lowe04,berg01}. Past research has shown that, employing multiple descriptors rather than any single descriptor leads to better recognition~\cite{varma07,kumar07}. The problem of learning the optimal combination of the available descriptors for a particular classification task is studied. Multiple Kernel Learning (MKL) framework has been developed for learning an optimal combination of descriptors for object categorization. Existing MKL formulations often employ block $l$-1 norm regularization which is equivalent to selecting a single kernel from a library of kernels~\cite{LaCrBaGhMi04,bach04,son06,AlFrStYv08,XuJiKiLy08}. Since essentially a single descriptor is selected, the existing formulations maybe suboptimal for object categorization. A MKL formulation based on block $l$-$\infty$ norm regularization has been developed, which chooses an optimal combination of kernels as opposed to selecting a single kernel. A Composite Multiple Kernel Learning(CKL) formulation based on mixed $l$-$\infty$ and $l$-1 norm regularization has been developed. These formulations end in Second Order Cone Programs(SOCP). Other efficient alternative algorithms for these formulation have been implemented. Empirical results on benchmark datasets show significant improvement using these new MKL formulations. 
\end{abstract}
\section{INTRODUCTION}
Object Categorization is a challenging problem, especially when the images have clutter background, occlusions or different lighting conditions. In the past, many descriptors have been proposed which aid object categorization even in such adverse conditions. Each descriptor has its own merits and de-merits. Some descriptors are invariant to transformations while the others are more discriminative. For example, Scale Invariant Feature Transformation~(SIFT~\cite{lowe04}) is invariant to affine transformations, geometric blur descriptor~\cite{berg01} is robust to shape deformation and pyramid histogram of gradient~\cite{bosch07} is invariant to geometric and photometric transformations. Past research has shown that, employing multiple descriptors rather than any single descriptor leads to better recognition~\cite{varma07,kumar07}. The project focuses on the problem of learning the optimal combination of the available descriptors for a particular classification task.

AdaBoost for combining the descriptors has been developed which is inspired by the MKL work where each kernel is formed with different descriptors. Difference between AdaBoost SVM with different descriptors and MKL is AdaBoost gives weight on the SVM classifier, each SVM with different descriptors in kernel, where MKL gives weights on each kernel.

In~\cite{varma07,nils08,lin07,kumar07}, the authors employ the Multiple Kernel Learning (MKL) framework~\cite{LaCrBaGhMi04} to find the optimal combination of descriptors (kernels). The goal of MKL is to simultaneously optimize the combination of kernels and the usual classification objective. Most of the existing MKL formulations perform a $l$-1 regularization~\cite{AlFrStYv08,varma07} over the kernels. This is equivalent to selecting the best kernel from the given set of kernels; which, as discussed earlier, might be suboptimal for object categorization tasks. One way to circumvent this problem of optimal weights being zero for many of the kernels, was introduced in~\cite{varma07}, where an additional constraint to employ prior information is included.

A new formulation for the MKL problem based on block $l$-$\infty$ and mixed norm($l$-$\infty$ and $l$-1 norm)regularization has been developed. It is well known that such a regularization would induce ``equal weightage'' to all the kernels rather than sparsity is developed. Hence would be ideal for applications such as object categorization, in which a combination of the descriptors is known to perform better than any single descriptor. 

These new MKL formulations are Second Order Cone Programs(SOCP) which can be solved using solvers like Mosek, SeDuMi, etc. Other efficient alternative algorithms are also developed which alternates between SVM optimization parameter and kernel weights. Empirical results on Caltech-4, Caltech-101 and Oxford flower datasets show significant improvement using these new MKL formulations.

The outline of the report is as follows: section~\ref{sec:rel} briefly reviews the work on object recognition. Existing MKL formulations is given section~\ref{sec:mkl} and the new MKL formulations, is presented in section~\ref{sec:nmkl}~\ref{sec:cmkl}. In the subsequent section, efficient algorithms for solving the proposed MKL formulation are discussed. Section~\ref{sec:exp} presents experimental results on synthetic and real-world datasets which illustrate the merits of the new MKL formulation. The results show that the new formulations achieves better recognition compared to state-of-the-art, which is an $l$-1 regularization based formulation. Video change detection problem is presented in section~\ref{sec:vid}. The report concludes in section~\ref{sec:conc} by summarizing the work.

\section{Related Work}\label{sec:rel}
This section provides some of work done in area of Machine learning involved in Object categorization. SVM-KNN~\cite{1153559} gets motivation from Local learning which uses K-Nearest neighbor to select local training point and uses SVM algorithm in those local training points for classification of object. Main problem here is time taken for classification. 

Multiple kernel learning considers the scenario where several descriptors (kernels) for a particular classification task are available. It aims to simultaneously learn the optimal combination of the given kernels and the optimal classifier parameters that maximize the generalization ability. Most of the work on MKL, since it was first introduced in~\cite{LaCrBaGhMi04}, concentrates on the employment of a block $l$-1 regularization. The main features of it being: a) $l$-1 regularization leads to sparse combination of the kernels, and hence automatically performs feature selection b) very efficient algorithms to solve the formulation exist~\cite{bach04,son06,AlFrStYv08,XuJiKiLy08}.

There has been lot of work on combining descriptors for the object categorization task~\cite{varma07,nils08,bosch07,kumar07,lin07,zhang07,dg1,dg2,dg3,dg4,dg5,dg6}. In~\cite{bosch07}, the authors introduce spatial pyramid kernel and combine shape (pyramid histogram of gradient), appearance descriptors for object classification. In~\cite{kumar07}, the Support Kernel Machine~\cite{bach04}, which is again based on $l$-1 regularization, is employed for combining descriptors for object categorization. In~\cite{lin07}, a sample dependent local ensemble kernel machine is learned for object categorization. In~\cite{varma07}, the authors use six descriptors for object categorization and employ a MKL formulation for learning the optimal combination of descriptors. However, as observed by the authors, most of the (important) kernels get eliminated in the optimal combination. This, as discussed above, is a consequence of employing the $l$-1 regularization. In order to circumvent the problem of optimal weights being zero for most of the kernels, the authors introduce additional constraints and parameters to utilize additional prior information regarding the kernels. This MKL formulation~\cite{varma07} is known to achieve state-of-the-art performance for many object recognition tasks. In~\cite{nils08}, four descriptors for flower classification task were combined using the multiple kernel learning formulation in~\cite{varma07} and this is shown to achieve state-of-the-art performance on such tasks. \\

In summary, most of the existing methodologies for object categorization employ the $l$-1 regularization based MKL formulation and its variants. As discussed earlier, such a regularization leads to kernel selection rather than kernel combination and hence suboptimal for object categorization tasks. MKL formulation with block $l$-$\infty$ regularization and CKL, which is more suited for combining kernels as opposed to selecting kernels is presented in this report.
\section{Multiple kernel learning}\label{sec:mkl}
This section gives brief introduction about MKL. Let $\bx^k_i$ denote the feature vector of the $i^{th}$ training datapoint in the $k^{th}$ feature space ($k^{th}$ kernel). Suppose $y_i$ denotes its label. Let $\bX_{k}$ represent the matrix whose columns are the training datapoints in the $k^{th}$ feature space. Also, let $\bK_k\equiv\bX_k^\top\bX_k$, be the gram matrix of the training datapoints in the $k^{th}$ feature space. Note that $\bX_k$ may not be explicitly known; the gram matrices, $\bK_k$, are assumed to be known. Let $\by, \bY$ represent the column vector, diagonal matrix with entries as labels of the training datapoints respectively. \\
Let the discriminating hyperplane be $\sum_{k=1}^l \bw_k^\top\bx^k - b = 0$ (here, $\bx^k$ denotes the $k^{th}$ feature space representation of the datapoint, $\bx$. $l$ is the number of kernels given). The usual soft-margin Support Vector Machine (SVM)~\cite{Vapnik98,dg3} formulation with this notation is: 
\begin{eqnarray*}
\min_{\bw_k,b,\xi_i} & \frac{1}{2}\left[\sum_{k=1}^l \|\bw_k\|_2^2\right] + C\sum_i\xi_i&\\
\textup{s.t.} & y_i\left(\sum_{k=1}^l \bw_k^\top\bx_i^k - b\right) \ge 1-\xi_i,&\\
&  \xi_i\ge0,\ \forall\ i=1,\ldots,m & \textup{\textbf{(L2-MKL)}}
\end{eqnarray*}
This is evident by considering $\bw\equiv\left[\bw_1^\top \ldots \bw_l^\top\right]^\top$ and $\bx_i\equiv\left[\left(\bx_i^1\right)^\top\ldots\left(\bx_i^l\right)^\top\right]^\top$ in the usual SVM formulation. It is easy to see that in this case the gram-matrix of the datapoints $\bx_i,\ i=1,\ldots,m$ ($m$ is the number of training datapoints) is nothing but $\sum_{k=1}^l\bK_k$. Hence, in the context of MKL, the optimal kernel with this formulation is nothing but a simple sum of the given kernels. \\
Another alternative which has been extensively explored in the past~\cite{LaCrBaGhMi04,AlFrStYv08} was to employ a block $l$-1 regularization in order to perform kernel selection. This formulation can be written as:
\begin{eqnarray*}
\min_{\bw_k,b,\xi_i} & \frac{1}{2}\left[\sum_{k=1}^l \|\bw_k\|_2\right]^2 + C\sum_i\xi_i&\\
\textup{s.t.} & y_i\left(\sum_{k=1}^l \bw_k^\top\bx_i^k - b\right) \ge 1-\xi_i,&\\
&  \xi_i\ge0,\ \forall\ i=1,\ldots,m & \textup{\textbf{(L1-MKL)}}
\end{eqnarray*}
Note that this formulation performs a $l$-1 regularization over $\|\bw_k\|_2,\ k=1,\ldots,l$. Hence it automatically performs kernel selection and is equivalent to selecting one (the best) of the given kernels. Since the formulation promotes sparsity in the usage of the given kernels, it is best suited for feature selection applications rather than for applications like object categorization where each kernel is believed to provide important information regarding the classification problem at hand. \\
\section{$L$-$\infty$ regularization MKL Formulation}\label{sec:nmkl}
The alternative to that of $l$-1 regularization is to perform block $l$-$\infty$ regularization. Such a regularization promotes the use of all the kernels while assuming they are equally preferable. The proposed MKL formulation can be written~as:
\begin{eqnarray*}
\min_{\bw_k,b,\xi_i} & \left[\max_{k=1}^l \|\bw_k\|_2^2\right] + C\sum_i\xi_i&\\
\textup{s.t.} & y_i\left(\sum_{k=1}^l \bw_k^\top\bx_i^k - b\right) \ge 1-\xi_i,&\\
&  \xi_i\ge0,\ \forall\ i=1,\ldots,m & \textup{\textbf{(Li-MKL)}}
\end{eqnarray*}
In the remainder of this section, the ranges of the indices $i,k$ are omitted for convenience. The \textbf{(Li-MKL)} formulation is same as:
\begin{eqnarray*}
 \min_{t,\bw_k,b,\xi_i} & t + C\sum_i\xi_i\\
\textup{s.t.} & y_i(\sum_{k} \bw_k^\top\bx_i^k - b) \ge 1-\xi_i,\ \xi_i\ge0,\ \|\bw_k\|_2^2\le t
\end{eqnarray*}

In the following text the dual of the proposed MKL formulation is derived. The Lagrangian turns out to be:
$$\mathcal{L}=t + C\sum_i\xi_i + \sum_i\alpha_i\left(1-\xi_i-y_i\left(\sum_k \bw_k^\top\bx_i^k - b\right)\right) -$$ $$\sum_i\beta_i\xi_i+\sum_k\lambda_k\left(\|\bw_k\|_2^2-t\right)$$
where $\alpha_i, \beta_i, \lambda_k \ge 0$ are the Lagrange multipliers. From the KKT conditions:
\begin{align}\label{eqn:kkt1}
 \nabla_{\bw_k}\mathcal{L}=0 &\Rightarrow \lambda_k\bw_k = \frac{1}{2}\sum_i\alpha_iy_i\bx_i^k\\
\frac{\partial \mathcal{L}}{\partial t} = 0 &\Rightarrow \sum_k\lambda_k=1\\
\frac{\partial \mathcal{L}}{\partial b} = 0 &\Rightarrow \sum_i\alpha_iy_i=0\\
\frac{\partial \mathcal{L}}{\partial \xi_i} = 0 &\Rightarrow C=\alpha_i+\beta_i
\end{align}
Now, suppose that all the gram-matrices $K_k$ are positive-definite (add a small ridge if singular, see also~\cite{AlFrStYv08}). Then, (\ref{eqn:kkt1}) implies that if $\lambda_k = 0$ for some $k$, then $\alpha_i=0,\ \forall\ i$. Clearly, in this case rest of the $\lambda_k$ must also be zero --- which is not possible since $\sum_k\lambda_k=1$. Hence $\lambda_k>0\ \forall\ k$.

Eliminating the primal variables, the dual can be written as:
\begin{eqnarray}\label{eqn:dual}
 \min_{\balpha,\blambda} & \frac{1}{2}\balpha^\top\bQ(\blambda)\balpha - \bone^\top\balpha \nonumber\\
\textup{s.t.} & \bzero \le \balpha \le C\bone, \by^\top\balpha=0, \nonumber\\
& \blambda\ge\bzero, \bone^\top\blambda=1
\end{eqnarray}
where $\bQ(\blambda)\equiv\frac{1}{2}\sum_k\frac{\bY\bK_k\bY}{\lambda_k}$ and $\blambda, \balpha$ denote the column vectors with entries as $\lambda_k, \alpha_i$ respectively. 

Though the dual in (\ref{eqn:dual}) has more variables, it gives more insight into the structure of the solution. Consider re-writing the dual (\ref{eqn:dual}) in the following way:
\begin{eqnarray}\label{eqn:split1}
 \min_{\blambda} & J(\blambda)\nonumber\\
\textup{s.t.} & \blambda\ge\bzero, \bone^\top\blambda=1
\end{eqnarray}
where $J(\blambda)$ is the optimal value of the following convex QP:
\begin{align}\label{eqn:split2}
J(\blambda) \equiv & \min_{\balpha} \ \ \ \frac{1}{2}\balpha^\top\bQ(\blambda)\balpha - \bone^\top\balpha \nonumber\\
&\textup{s.t.} \ \ \ \  \bzero \le \balpha \le C\bone, \by^\top\balpha=0
\end{align}

Note that (\ref{eqn:split2}) is nothing but a usual SVM problem and hence the optimal $\alpha$ is very sparse. Infact, algorithms which exploit this sparsity in solution and outperform standard QP solvers exist~\cite{Platt99}.
\subsection{Algorithms for solving the Li-MKL Formulation}\label{sec:alg}
The \textbf{Li-MKL} formulation, can be solved using standard Second Order Cone Program (SOCP) solvers(e.g., \texttt{SeDuMi}\footnote{\url{http://sedumi.ie.lehigh.edu/}}, \texttt{Mosek}\footnote{\url{www.mosek.com}}). However the optimization problem would involve $l$ conic quadratic constraints ($l$, the number of kernels, can be large). Also, the size of the optimization problem ($m$, the number of training datapoints), can be large. Hence generic cone solvers fail to solve for large $l$ or $m$. Interestingly, there  are more efficient ways to solve the dual formulations (\ref{eqn:dual}). The following sections explain in brief the possible methodologies.
\subsection{Alternating Minimization Algorithm}
The dual (\ref{eqn:dual}) can be solved efficiently using an alternating minimization algorithm in the variables $\balpha$ and $\blambda$. Note that for a fixed value of $\blambda$, (\ref{eqn:dual}) is nothing but the SVM dual (which has very efficient scalable solvers). Also, for fixed value of $\balpha$, the minimization wrt. $\blambda$ is the following simple problem:
\begin{eqnarray}\label{eqn:sp}
 \min_{\lambda_k} & \sum_k\frac{D_k}{\lambda_k} \nonumber\\
\textup{s.t.} & \lambda_k\ge0, \sum_k\lambda_k=1
\end{eqnarray}
where $D_k\equiv\balpha^\top\bQ_k\balpha$. It is easy to show that the optimal values of $\lambda_k$ for the problem (\ref{eqn:sp}) are ($\lambda_k>0\ \forall\ k$):
\begin{equation}\label{eqn:uk}
 \lambda_k=\frac{\sqrt{D_k}}{\sum_k\sqrt{D_k}}
\end{equation}

Hence the following iterative algorithm can be employed to solve (\ref{eqn:dual}) efficiently:
\begin{enumerate}
 \item Initialize with $\lambda_k^{(0)}=\frac{1}{l}$ where $l$ is the number of kernels.
\item At iteration $i\ge1$, solve a standard SVM problem with Hessian as $\bQ(\blambda^{(i-1)})$ for $\balpha^{(i)}$.
\item Using $\balpha^{(i)}$ compute $D_k^{(i)}$. Update $\blambda^{(i)}$ using (\ref{eqn:uk}).
\item repeat until, say, change in objective value of (\ref{eqn:dual}) is negligible.
\end{enumerate}
\section{Composite MKL Formulation}\label{sec:cmkl}
This section explains Composite MKL. Suppose $n$ descriptors are available. Further, for each of these descriptors Kernels (linear, polynomial, Gaussian) are defined. Let number of Kernels of the $j^{th}$ descriptor be denoted by $n_j$. Also, let $\phi_{jk}$ denote the mapping induced by the $k^{th}$ Kernel of the $j^{th}$ descriptor. The hyperplane classifier to be learnt has the form $\sum_{j=1}^n\sum_{k=1}^{n_j}\bw_{jk}^\top\phi_{jk}(\bx)-b=0$. The objective is to choose the ``best'' combination of these Kernels in order to maximize the generalization. The idea is to combine the Kernels in such a way that: a) all descriptors are given equal priority (weightage) b) best of the Kernels in each descriptor are selected. In other words, perform an $l$-$\infty$ regularization over the parameters ($\bw_{jk}$) such that each descriptor is given equal priority. Further, perform an $l$-1 regularization such that sparsity in selection of Kernels belonging to each descriptor is encouraged. Mathematically, the formulation can be written as:
\begin{eqnarray}
 \min_{\bw_{jk},b,\xi_i} & \left[\max_j\left(\sum_{k=1}^{n_j}\|\bw_{jk}\|_2\right)^2\right] + C\sum_i\xi_i\nonumber\\
\textup{s.t.} & y_i\left(\sum_{j=1}^n\sum_{k=1}^{n_j}\bw_{jk}^\top\phi_{jk}(\bx_i)-b\right)\\
\ge1-\xi_i\ \xi_i\ge0 & \textup{\textbf{(CKL)}}
\end{eqnarray}
where $\{(\bx_i,y_i), i=1,\ldots,m\}$ is the training dataset. $C$ is the regularization parameter.

Let $\by$ denote the vector with entries as the labels. Let $S_{m}$ be the set $\{\balpha\ \in\ \RR^m\ | \ 0\le\balpha\le C,\ \by^\top\balpha=0 \}$. Denote the set $\{\bgamma\ \in \ \RR^n\ | \ \bgamma\ge0,\ \bone^\top\bgamma=1 \}$ by $\Delta_n$. The dual of the above formulation can be written as:
\begin{equation}\label{eqn:dual}
\min_{\blambda_j\in\Delta_{n_j}} \ \max_{\balpha\in S_m,\ \gamma\in\Delta_n} \ \bone^\top\balpha - \frac{1}{4}\balpha^\top\left[\sum_{j=1}^n\left(\frac{\sum_{k=1}^{n_j}\lambda_{jk}\bQ_{jk}}{\gamma_j}\right)\right]\balpha
\end{equation}
where $\bQ_{jk}\equiv\bY\bK_{jk}\bY$. Here, $\bK_{jk}$ is the gram-matrix of the training datapoints with the $k^{th}$ Kernel of the $j^{th}$ descriptor and $\bY$ is the diagonal matrix with entries as the labels. The gram-matrices are assumed to be positive definite.

One can solve the above dual~(\ref{eqn:dual}) using a simple alternating algorithm described below. Due to compactness of feasibility sets and convexity of the objective, the order of min., max. can be rearranged. Also since the variables $\blambda_{j_1}$ are not inter-linked with the variables $\blambda_{j_2}$, for $j_1\ne j_2$, instead of minimizing sum over $j$ index one can sum the minima:
$$\equiv \max_{\balpha\in S_m} \max_{\bgamma\in\Delta_n} \min_{\blambda_j\in\Delta_{n_j}} \bone^\top\balpha -
\frac{1}{4}\balpha^\top\left[\sum_{j=1}^n\left(\frac{\sum_{k=1}^{n_j}\lambda_{jk}\bQ_{jk}}{\gamma_j}\right)\right]\balpha$$
$$\equiv \max_{\balpha\in S_m} \bone^\top\balpha - \frac{1}{4} \min_{\bgamma\in\Delta_n} \sum_{j=1}^{n} 
\frac{\max_{\blambda_j\in\Delta_{n_j}} \ \sum_{k=1}^{n_j}\balpha^\top\lambda_{jk}\bQ_{jk}\balpha}{\gamma_j}$$

Now it is easy to see that for fixed values of $\bgamma,\blambda_j\ \forall\ j$, the problem wrt. $\balpha$ is same as the SVM problem. Also, for fixed values of $\balpha$, the problem wrt. $\bgamma,\blambda_j\ \forall\ j$ is the following simple problem:
\begin{equation}\label{eqn:small}
\min_{\bgamma\in\Delta_n} \sum_{j=1}^{n} \frac{\max_{\blambda_j\in\Delta_{n_j}} \ \sum_{k=1}^{n_j}\balpha^\top\lambda_{jk}\bQ_{jk}\balpha}{\gamma_j}
\end{equation}
which has a closed form solution described below. Consider solving $$\max_{\blambda_j\in\Delta_{n_j}} \ \sum_{k=1}^{n_j}\balpha^\top\lambda_{jk}\bQ_{jk}\balpha$$ for a particular $j$. This amounts to just picking the maximum among $\balpha^\top\bQ_{jk}\balpha$ for $k=1,\ldots,n_j$. Let these maxima be denoted by $D_j (\ge0)$. Hence (\ref{eqn:small}) is equivalent to the following problem:
$$\min_{\bgamma\in\Delta_n} \sum_{j=1}^n\frac{D_j}{\gamma_j}$$
The optimal solution for this problem is: $\gamma_j=\frac{\sqrt{D_j}}{\sum_j\sqrt{D_j}}$.

The overall algorithm is as follows:
\begin{itemize}
\item Initialize with $\bgamma^{(0)}=\frac{1}{n}$ and $\blambda_j^{(0)}=\frac{1}{n_j}$.
\item At iteration $i\ge1$, solve an SVM taking kernel as $\sum_{j=1}^n\frac{\sum_{k=1}^{n_j}\lambda_{jk}^{(i-1)}\bQ_{jk}}{\gamma_j^{(i-1)}}$. Update $\balpha^{(i)}$ as the solution of this SVM.
\item Using updated values of $\balpha^{(i)}$, compute the closed form solution of (\ref{eqn:small}) using the methodology described above.
\item Repeat until convergence.
\end{itemize}
\section{Numerical Experiments}\label{sec:exp}
This section presents the experimental results on standard object categorization. Various experiments also conducted using the Adaboost for combining descriptors on standard Object categorization dataset. The key idea is to show that the proposed $l$-$\infty$ regularization and Composite regularization based MKL formulation leads to better generalization than the $l$-1 regularization based MKL formulations, which represent state-of-the-art methodologies for object recognition. The results on synthetic and real-world data are summarized in sections~\ref{sec:syn}~and~\ref{sec:real} respectively. In all cases, the parameters for the respective methods were tuned on a validation set. Also, the accuracies reported are on unseen testsets and hence represent a true estimate of the generalization performance of the respective classifiers. All multi-class problems were handled using the one-vs-one scheme.
\subsection{Results using Adaboost}
Adaboost mentioned in the result is performed with following setup:
\begin{enumerate}
 \item Set of classifier for Adaboost is SVM.
 \item Each SVM in that set is build on different base kernel $K_i$ mentioned in previous section.
 \item Here each $K_i$ are build with descriptors like pyramid histograms of gradient, scale-invariant feature descriptors.
 \item Adaboost provides weight on the SVM classifier which is build on each such kernel mentioned above.
\end{enumerate}
This AdaBoost is inspired by the Multiple kernel learning work where each kernel is formed with different descriptors. Difference between AdaBoost with SVM(with different descriptors) and Multiple kernel learning is AdaBoost gives weight on the SVM classifier(each SVM with different descriptors in kernel) where in multiple kernel learning gives weights on each kernel in a SVM problem.
Following table shows result. \newline \newline
\begin{table}
\begin{tabular}{|l|l|l|}
\hline
Classifier & No of objects & Accuracy \\
\hline
NNC &  10 & 58\%\\ 
k-NNC & 10 & 62\% \\
SVM & 20 & 59\% \\
Local Learning & 10 & 76\% \\
AdaBoost + SVM & 10 & 78\% \\
AdaBoost + SVM & 20 & 63\% \\
\hline
\end{tabular}
\caption{Shows results on Caltech dataset with various classifiers and number of categories.}
\end{table}

\subsection{Results of $L$-$\infty$ MKL Experiments}
This section provides experimental results for the \textbf{Li-MKL}.
\subsubsection{Synthetic Data}\label{sec:syn}
In this section, results on synthetic datasets showing the benefit of the proposed methodology is presented. The key result to establish is that the \textbf{Li-MKL} formulation achieves better generalization, especially in cases where the redundancy in the given kernels is less, such as in applications like object categorization. For this, the experimental strategy given by~\cite{MaUlPaSo08}. We repeat the description of the experimental set-up here for the sake of completeness.

We wish to create $l$ kernels whose degree of redundancy is controlled by a single parameter $\rho$. First, $m$ datapoints are sampled from two independent Normal distributions with covariance as the identity matrix (dimensionality of data is $n$). Here, datapoints sampled from different Normals are assumed to belong to different classes. Now the features are grouped into $p$ disjoint sets ($p$ varies from 1 to $l$): $\bX_1,\ldots,\bX_p$ where $\bX_k\in\RR^{\frac{n}{p}\times m}$. We then sample $l-p$ copies from these disjoint sets, by randomly picking one by one from $\bX_1,\ldots,\bX_p$ with replacement. For each
of these $l$ sets randomly generate a linear transformation matrix $\bA_i\in\RR^{\tau\frac{n}{p}\times\frac{n}{p}}$ ($\tau$ is a parameter). The gram-matrices are computed as $\bK_k=\bX_k^\top\bA_k^\top\bA_k\bX_k$. Clearly, by varying $\rho=\frac{p}{l}$, the redundancy in the kernels can be varied. More specifically, $\rho=1$ represents the extreme case where the redundancy in kernels is zero, and hence represents the best-suited scenario for the proposed methodology. The other extreme case is $\rho=0$, where the redundancy is maximum, and hence an ideal scenario for employing the $l$-1 regularization based MKL.

\begin{figure}
\centering
\includegraphics[width=0.75\linewidth]{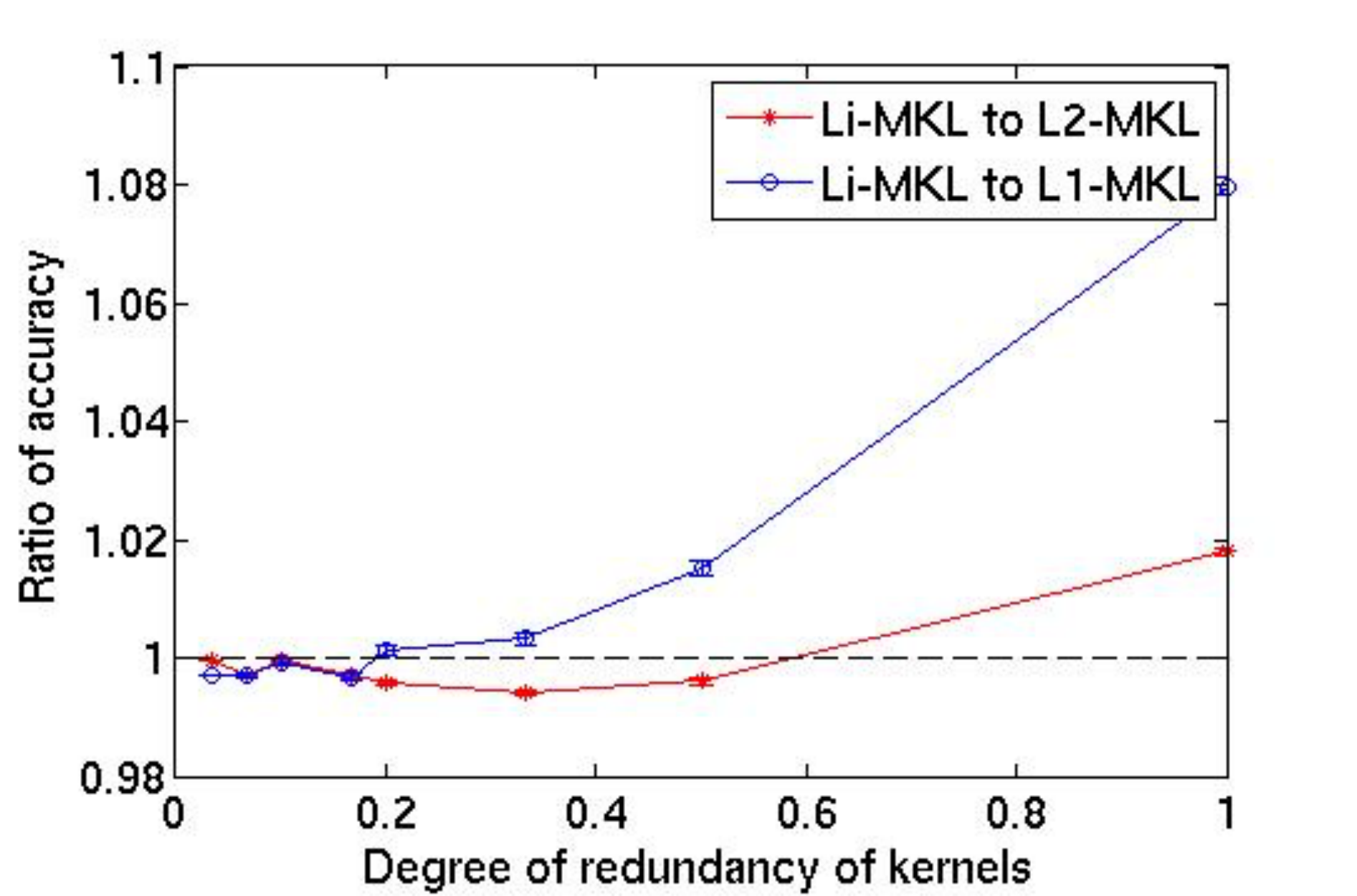}
\caption{Figure shows ratio of accuracy of \textbf{Li-MKL} to \textbf{L1-MKL} as a function of redundancy in kernels for synthetic data generated using the following parameters: $l=30$, $m=450$, $n=30$, $\tau=4$}\label{fig:syn}
\end{figure}

Figure~\ref{fig:syn} shows the plot of ratio of testset accuracies achieved by \textbf{Li-MKL} and \textbf{L1-MKL} vs. redundancy in the given kernels (vertical bars represent variance in accuracy). As a baseline for comparison, plot a similar graph for the ratio of testset accuracy achieved by \textbf{Li-MKL} and \textbf{L2-MKL}. Note that as the degree of redundancy decreases the ratio in case of both graphs increases; proving that \textbf{Li-MKL} is well-suited for applications like object categorization. In fact, observed a huge improvement in generalization over the \textbf{L1-MKL} when $\rho$ is near 1 (as high as $8\%$ and $2\%$ in case of \textbf{L2-MKL}). Also, in cases where the redundancy is high ($\rho$ is near 0), \textbf{Li-MKL} achieves generalization comparable to the other MKL formulations.
\subsubsection{Results on Caltech4 dataset}\label{sec:real}
\begin{figure}
\centering
\includegraphics[width=2.5in,height=2in]{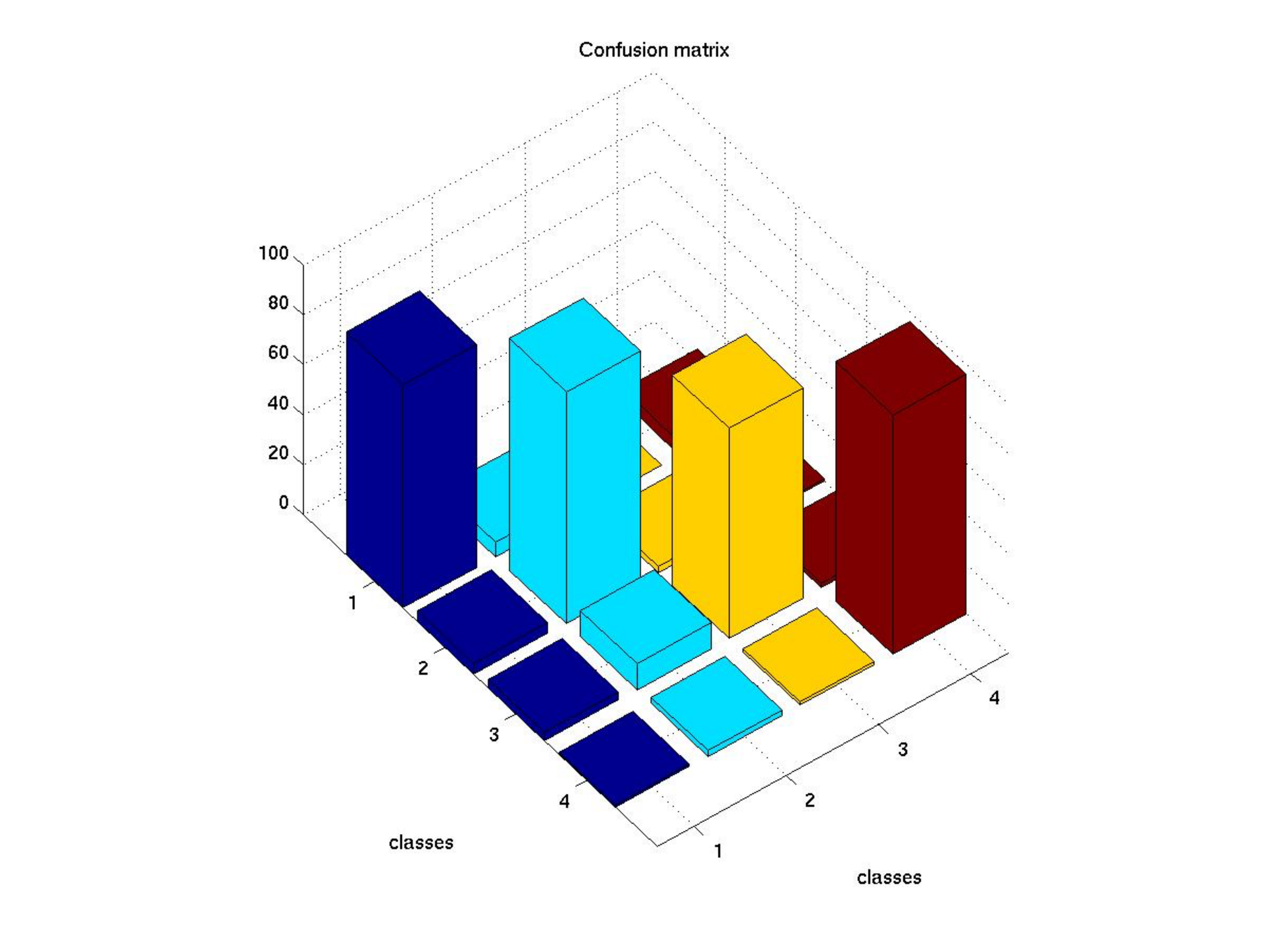}
\caption{Figure shows Confusion matrix on Caltech-4 dataset.}\label{fig:conf_4}
\end{figure}

\begin{figure}
\centering
\includegraphics[width=2in,height=2in]{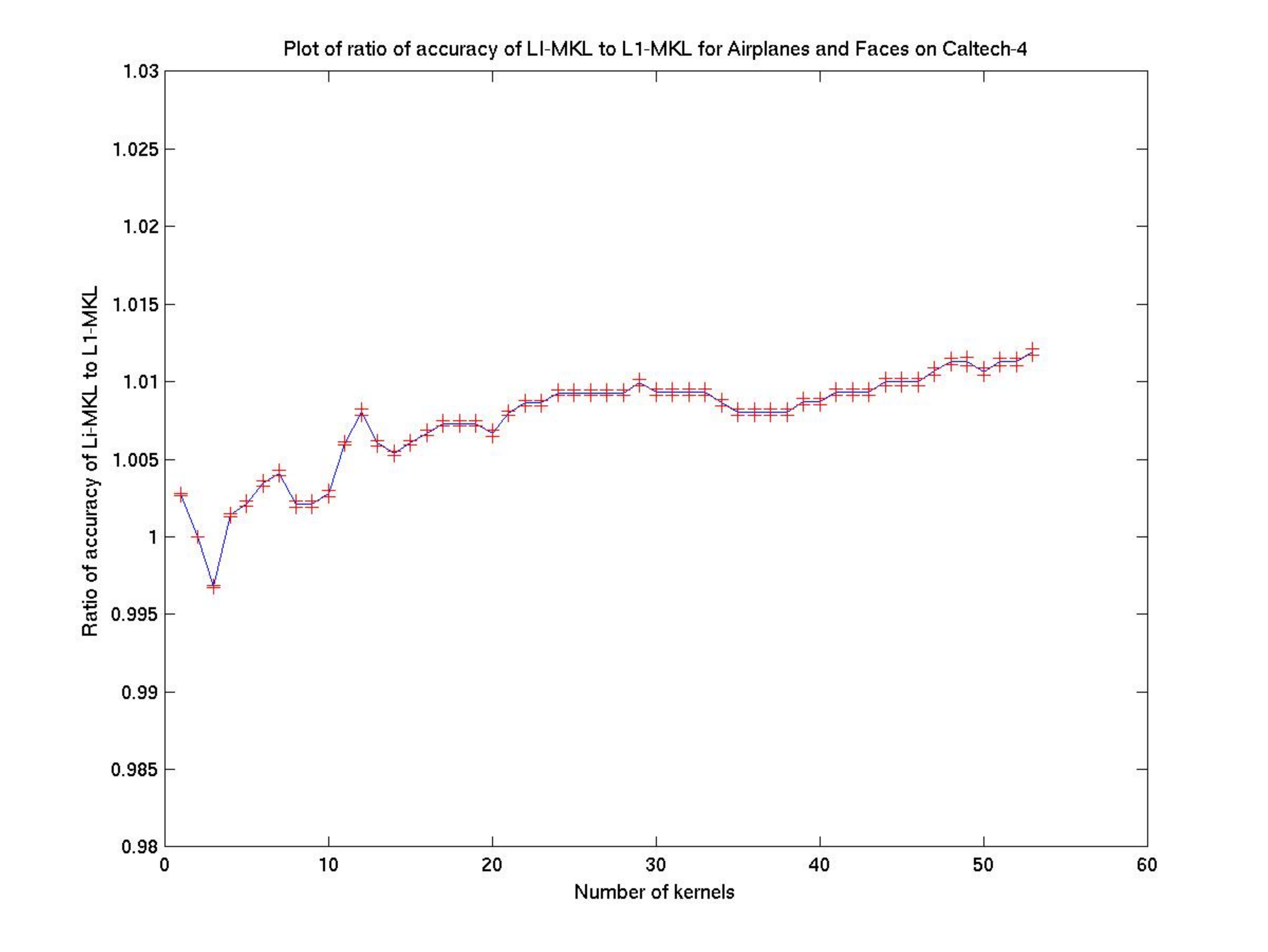}
\caption{Figure shows ratio of overall accuracies of \textbf{Li-MKL} to \textbf{L1-MKL} as function of number of kernels on Caltech-4 dataset(Airplane and Faces).}\label{fig:a_f}
\end{figure}

\begin{figure}
\centering
\includegraphics[width=2in,height=2in]{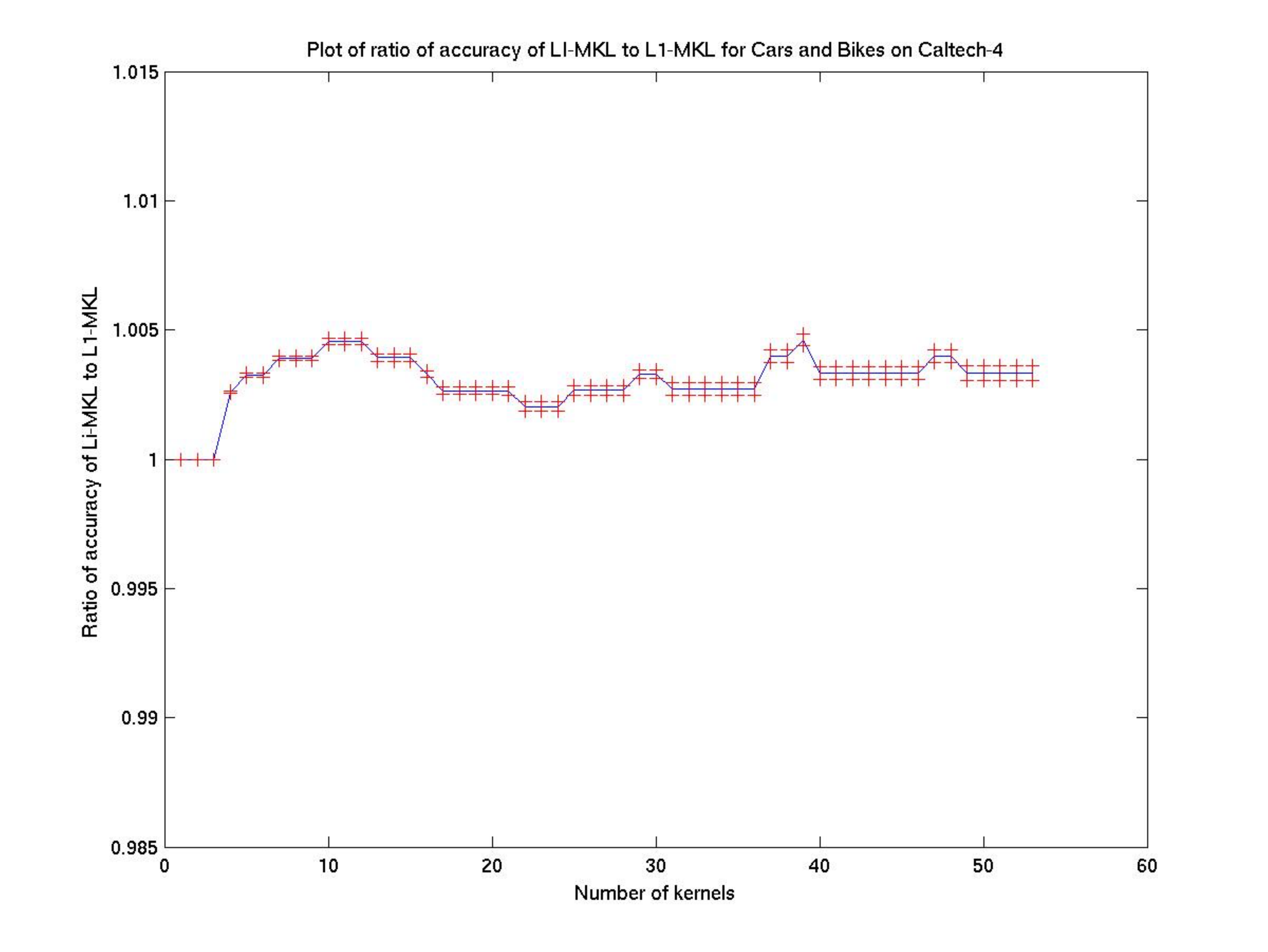}
\caption{Figure shows ratio of overall accuracies of \textbf{Li-MKL} to \textbf{L1-MKL} as function of number of kernels on Caltech-4 dataset(Bikes and Cars).}\label{fig:c_b}
\end{figure}

\begin{figure}
\centering
\includegraphics[width=2in,height=2in]{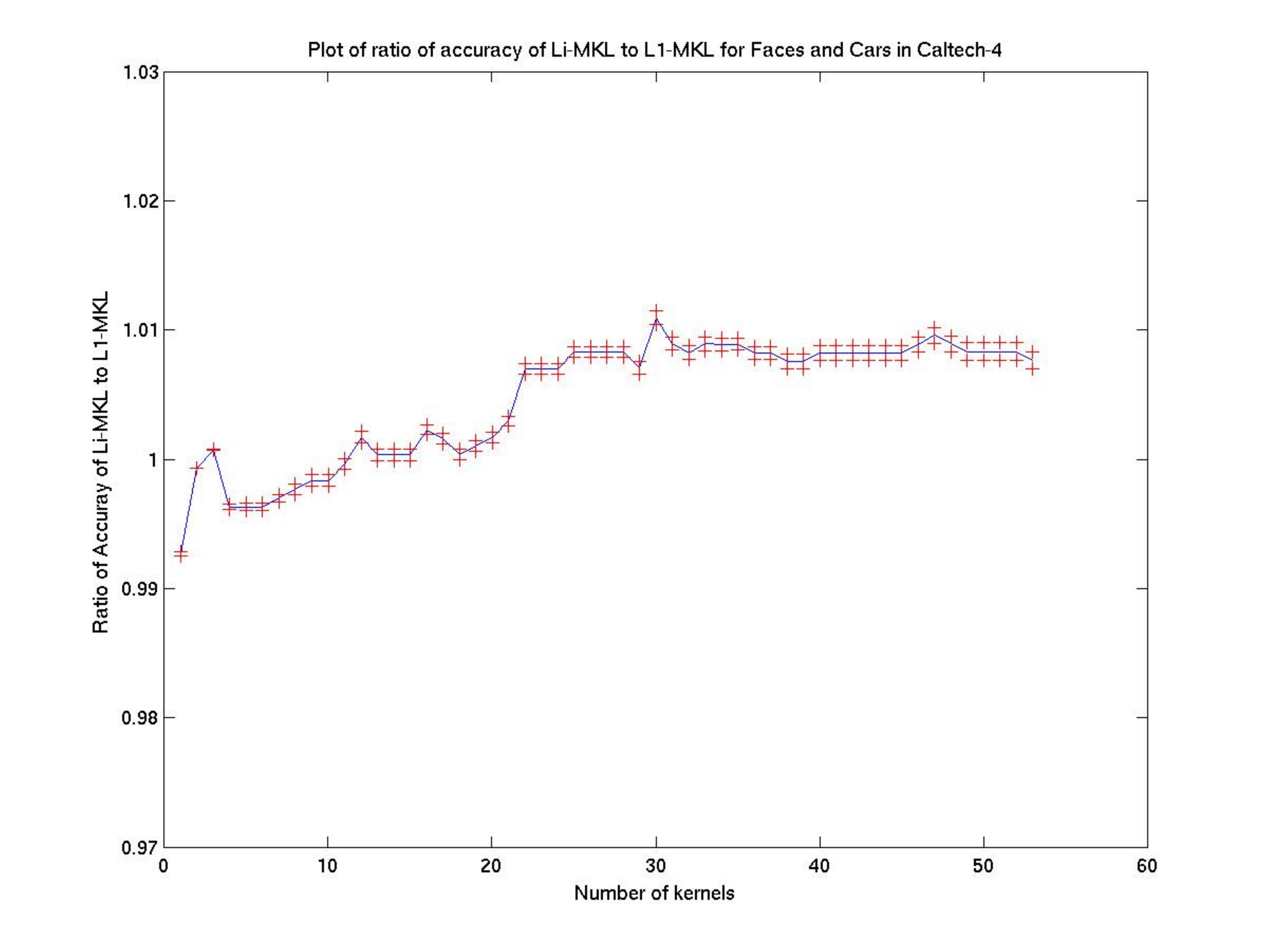}
\caption{Figure shows ratio of overall accuracies of \textbf{Li-MKL} to \textbf{L1-MKL} as function of number of kernels on Caltech-4 dataset(Cars and Faces).}\label{fig:c_f}
\end{figure}

\begin{figure}
\centering
\includegraphics[width=2in,height=2in]{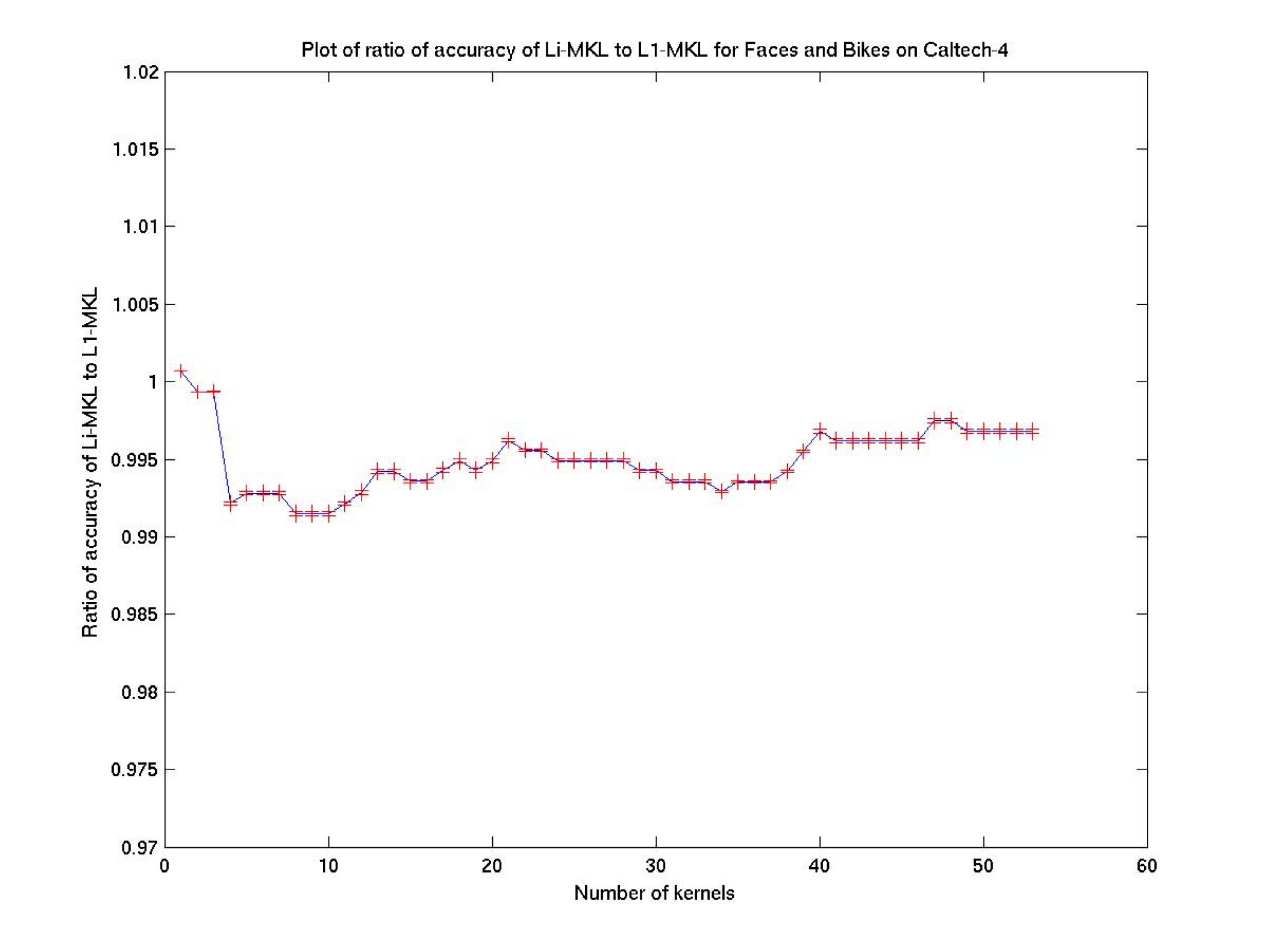}
\caption{Figure shows ratio of overall accuracies of \textbf{Li-MKL} to \textbf{L1-MKL} as function of number of kernels on Caltech-4 dataset(Faces and Bikes).}\label{fig:f_b}
\end{figure}

This section presents results on Caltech-4\footnote{\url{http://www.robots.ox.ac.uk/~vgg/data/data-cats.html}}. Caltech-4 dataset contains images of airplanes, cars, faces and bikes. We have taken 80 images for each class, of which 40 are randomly taken as the training/validation data and the remaining as test data. We have used Pyramid Histogram Of Gradient (PHOG) features generated\footnote{Code available at \url{http://www.robots.ox.ac.uk/~vgg/research/caltech/phog.html}} at various levels (1,2,3) and angles (180,360). We have generated kernels on these six PHOG features using different parameters for the polynomial and Gaussian kernel (9 for each feature, totally 54 kernels). This experimental procedure was repeated for 20 times with different training-test data splits. The mean testset accuracies obtained with \textbf{L1-MKL} and \textbf{Li-MKL} were 92.00$\pm$2.44\% and 93.50$\pm$2.14\% respectively. This shows that the \textbf{Li-MKL} achieves better generalization. Following figures~\ref{fig:ratio4}~\ref{fig:a_b}~\ref{fig:a_c}~\ref{fig:a_f}~\ref{fig:c_b}~\ref{fig:c_f}~\ref{fig:f_b} shows ratio of accuracies of \textbf{Li-MKL} to \textbf{L1-MKL} as function of number of kernels on Caltech-4 dataset. Figure~\ref{fig:conf_4} shows confusion for Caltech-4 dataset.

\begin{figure}
\centering
\includegraphics[width=2in,height=2in]{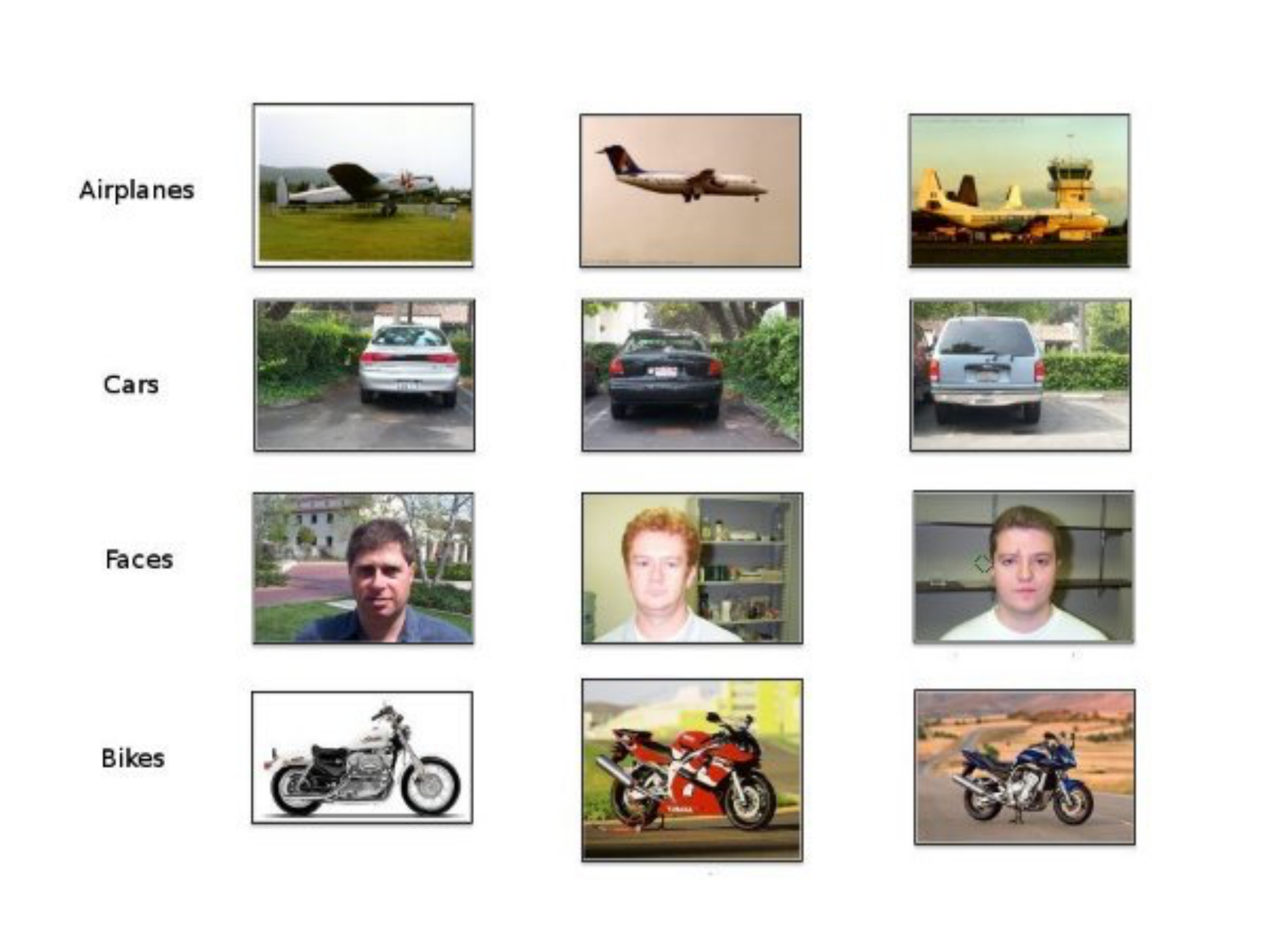}
\caption{Figure shows some example images from Caltech-4.}\label{fig:ex}
\end{figure}

\begin{figure}
\centering
\includegraphics[width=2.5in,height=2in]{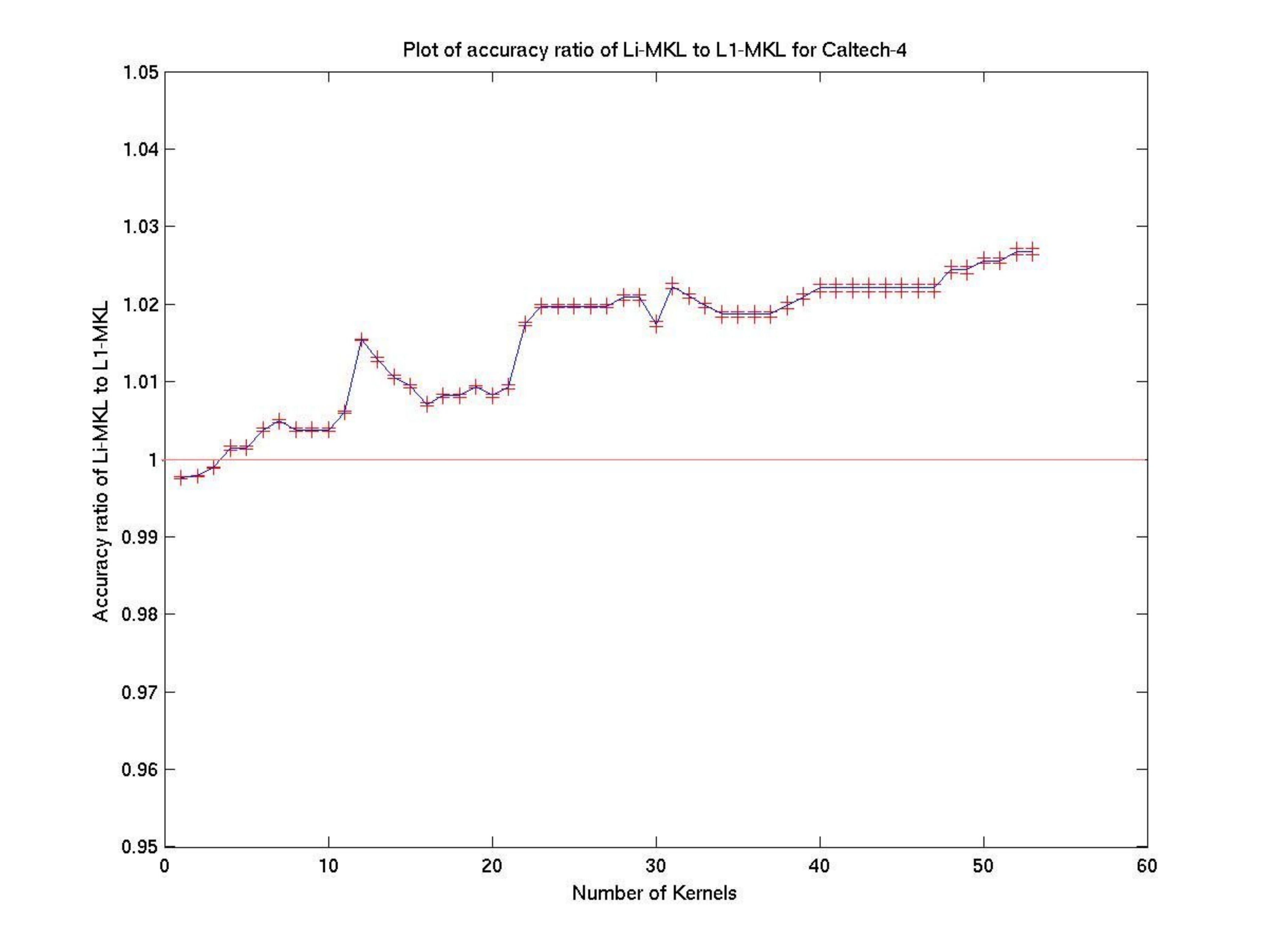}
\caption{Figure shows ratio of overall accuracies of \textbf{Li-MKL} to \textbf{L1-MKL} as function of number of kernels on Caltech-4 dataset.}\label{fig:ratio4}
\end{figure}

\begin{figure}
\centering
\includegraphics[width=2in,height=2in]{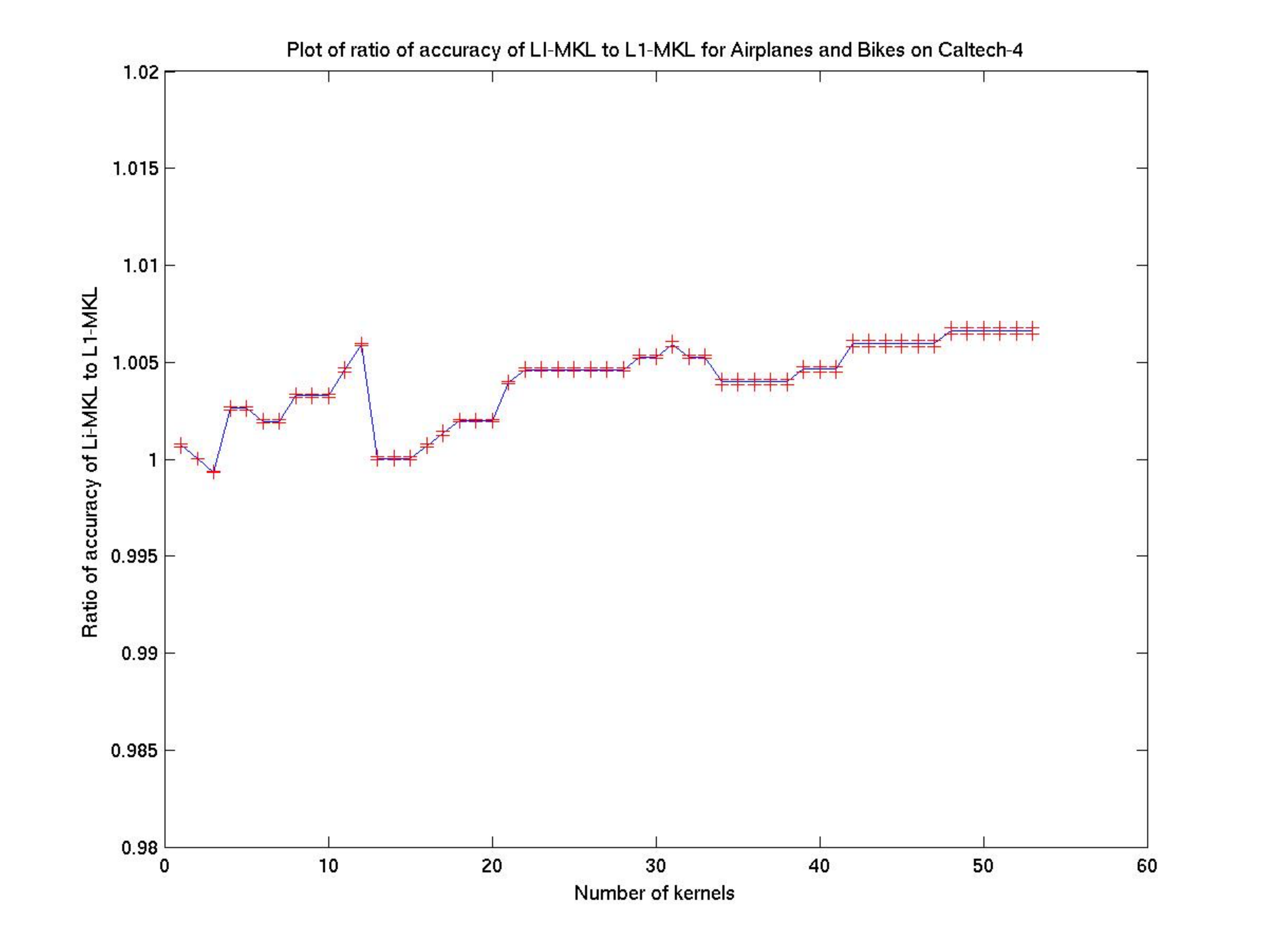}
\caption{Figure shows ratio of overall accuracies of \textbf{Li-MKL} to \textbf{L1-MKL} as function of number of kernels on Caltech-4 dataset(Airplane and Bikes).}\label{fig:a_b}
\end{figure}
\begin{figure}
\centering
\includegraphics[width=2in,height=2in]{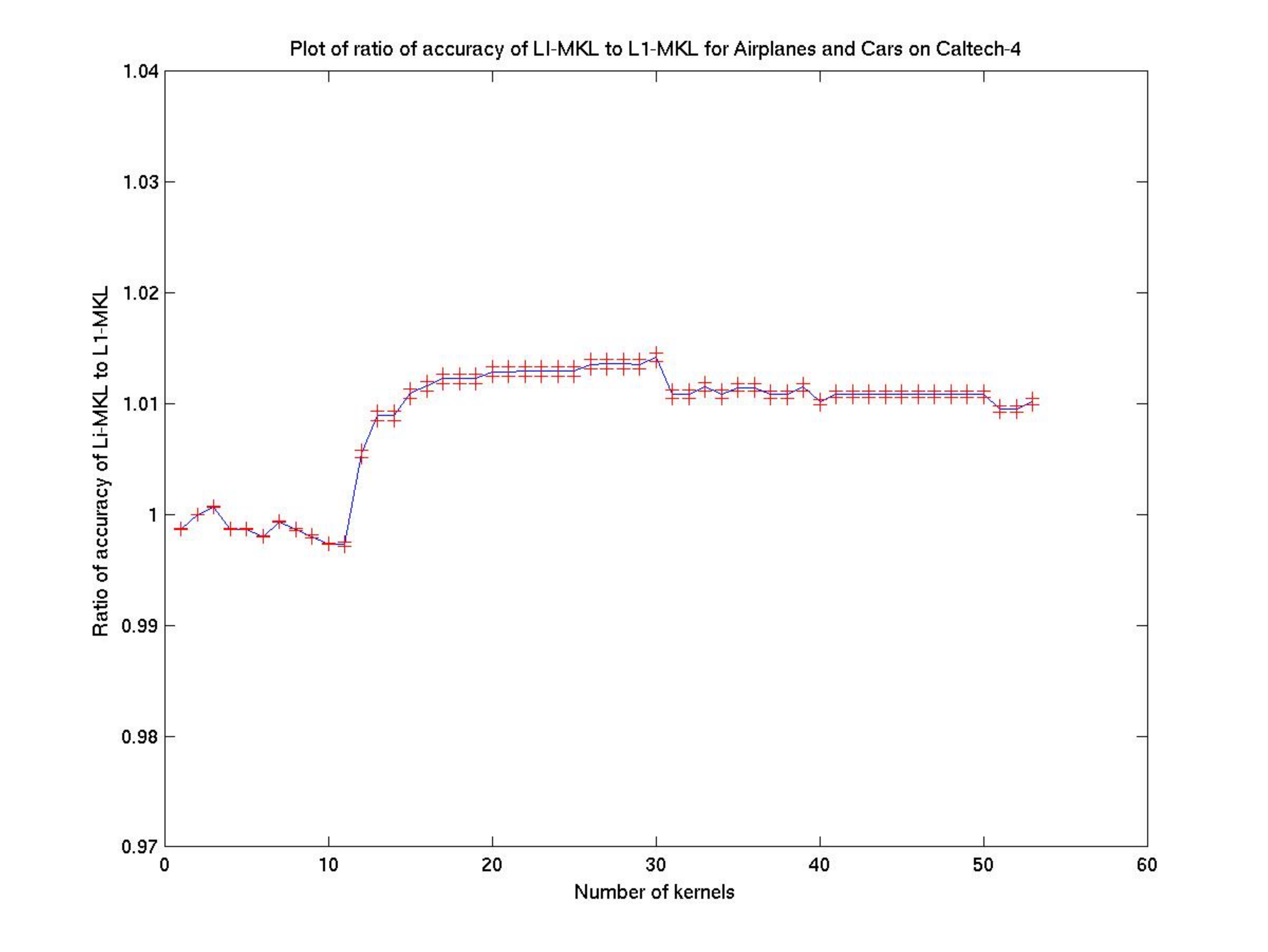}
\caption{Figure shows ratio of overall accuracies of \textbf{Li-MKL} to \textbf{L1-MKL} as function of number of kernels on Caltech-4 dataset(Airplanes and Cars).}\label{fig:a_c}
\end{figure}

\subsubsection{Results on Oxford dataset}\label{sec:real}
\begin{figure}
\centering
\includegraphics[width=2in,height=2in]{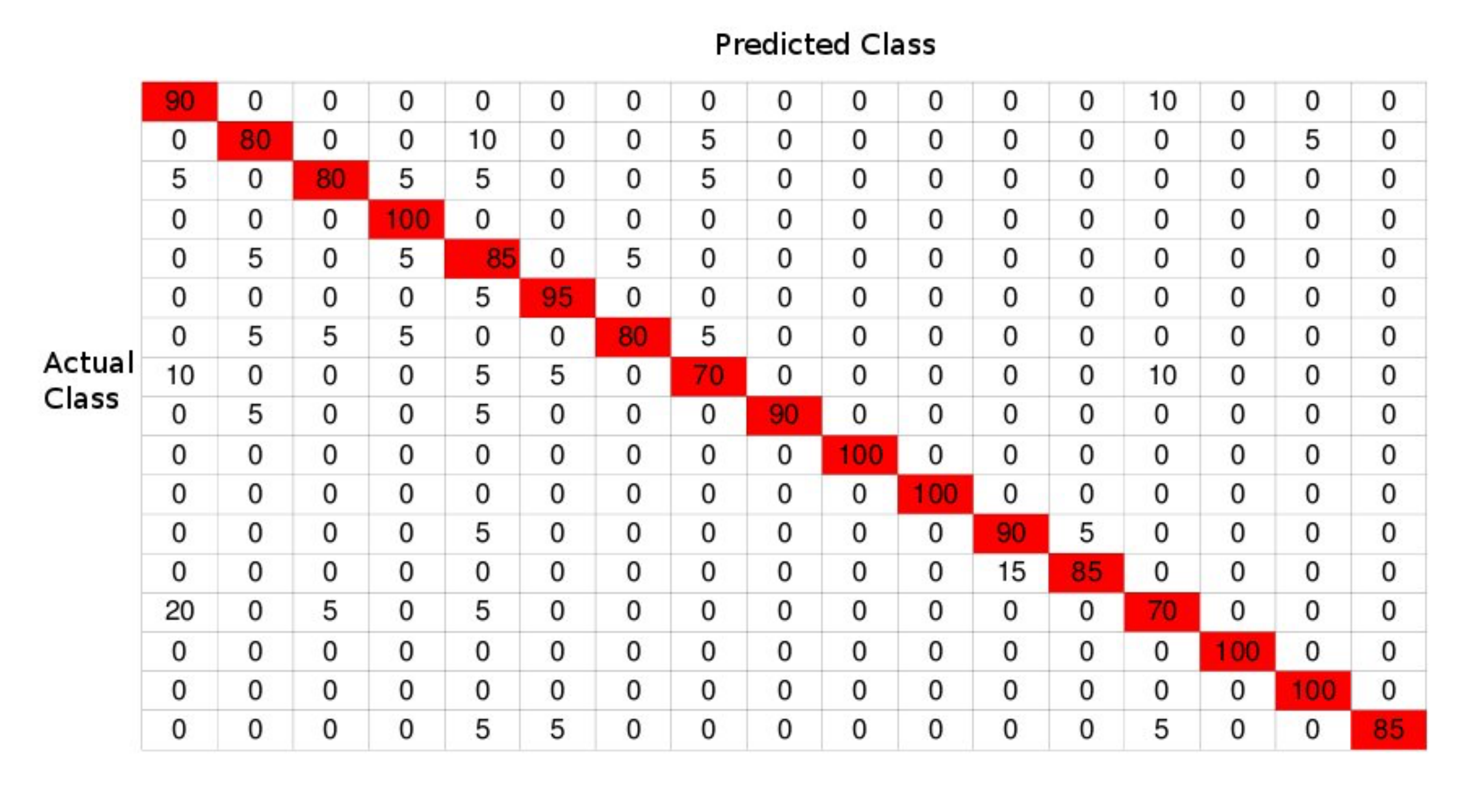}
\caption{Figure shows Confusion matrix on Oxford dataset.}\label{fig:conf_ox}
\end{figure}

\begin{figure}
\centering
\includegraphics[width=2in,height=2in]{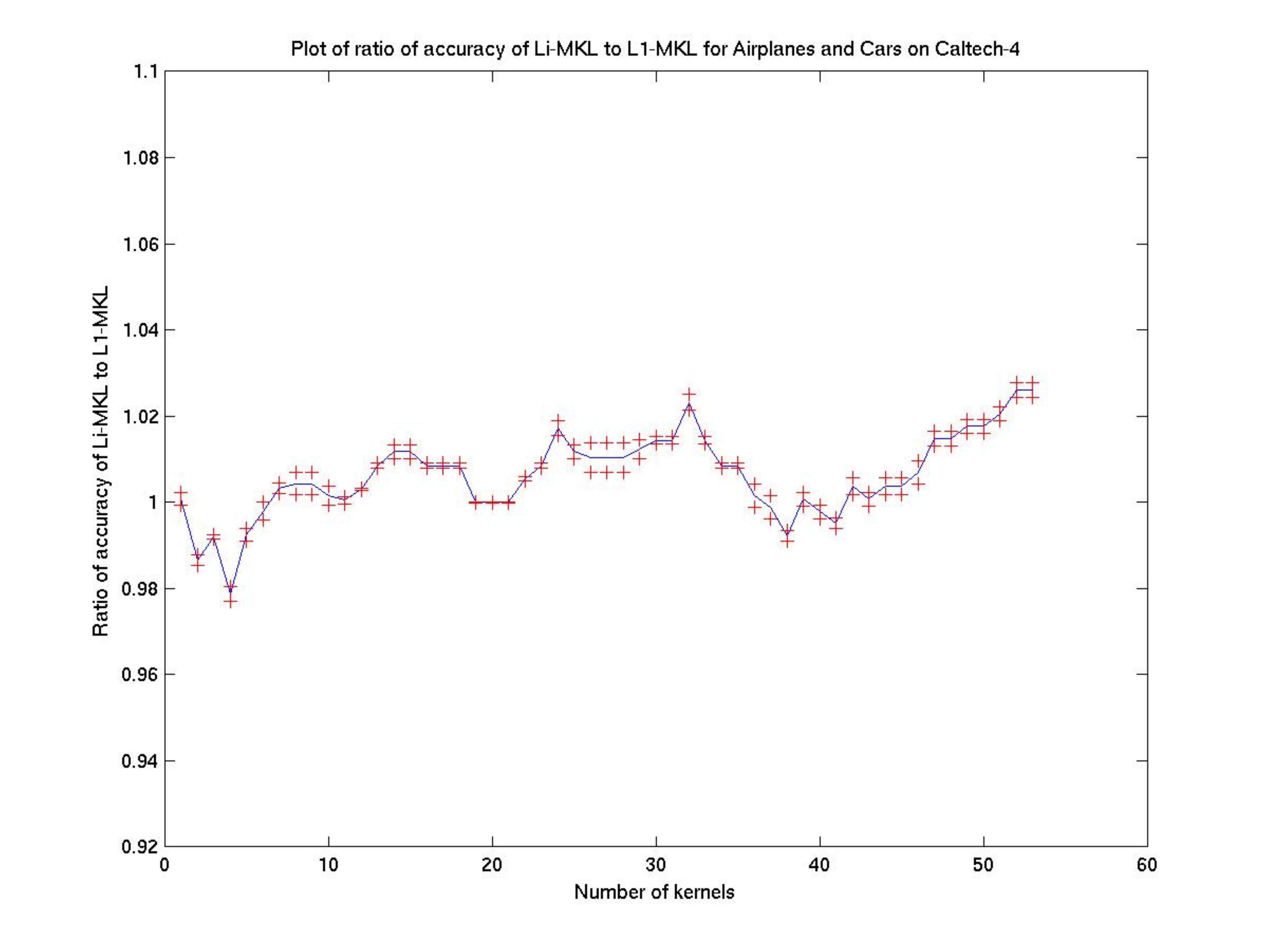}
\caption{Figure shows ratio of overall accuracies of \textbf{Li-MKL} to \textbf{L1-MKL} as function of number of kernels on some binary classification on Oxford dataset(Faces and Bikes).} \label{fig:ox_10_14}
\end{figure}

The task in the Oxford flower dataset is to categorize images of 17 varieties of flowers. This dataset contains 80 examples for each class. In~\cite{nils08}, the authors introduced four different features color, SIFT for foreground region, SIFT for foreground boundary, Histogram of Gradients for flowers. We have used the $\chi^2$ distances given in~\cite{nils08,nils06}\footnote{\url{http://www.robots.ox.ac.uk/~vgg/data/flowers/17/index.html}} for our experimentation on this dataset. We have used same training, validation and test splits as used in ~\cite{nils08}. The mean testset accuracy achieved by \textbf{L1-MKL} and \textbf{Li-MKL} are 85.88$\pm$1.83\% and 87.35$\pm$1.72\% respectively. Again, the results confirm that the proposed methodology achieves better generalization than state-of-the-art. The accuracy achieved by the proposed formulation is comparable to the best accuracy reported in~\cite{nils08}, which is 88.33$\pm$0.3\%. Note that this state-of-the-art accuracy was achieved after tuning the parameters for the various descriptors~\cite{nils08} and incorporating prior information following the strategy of~\cite{varma07}. As mentioned earlier, incorporating such prior information may further improve testset accuracies of the proposed formulation. Following figures~\ref{fig:ratio}~\ref{fig:ox_1_2}~\ref{fig:ox_2_5}~\ref{fig:ox_10_14} shows ratio of accuracies of \textbf{Li-MKL} to \textbf{L1-MKL} as function of number of kernels on Oxford flower dataset. Figure~\ref{fig:conf_ox} shows confusion for Oxford flower dataset.
\begin{figure}
\centering
\includegraphics[width=2in,height=2in]{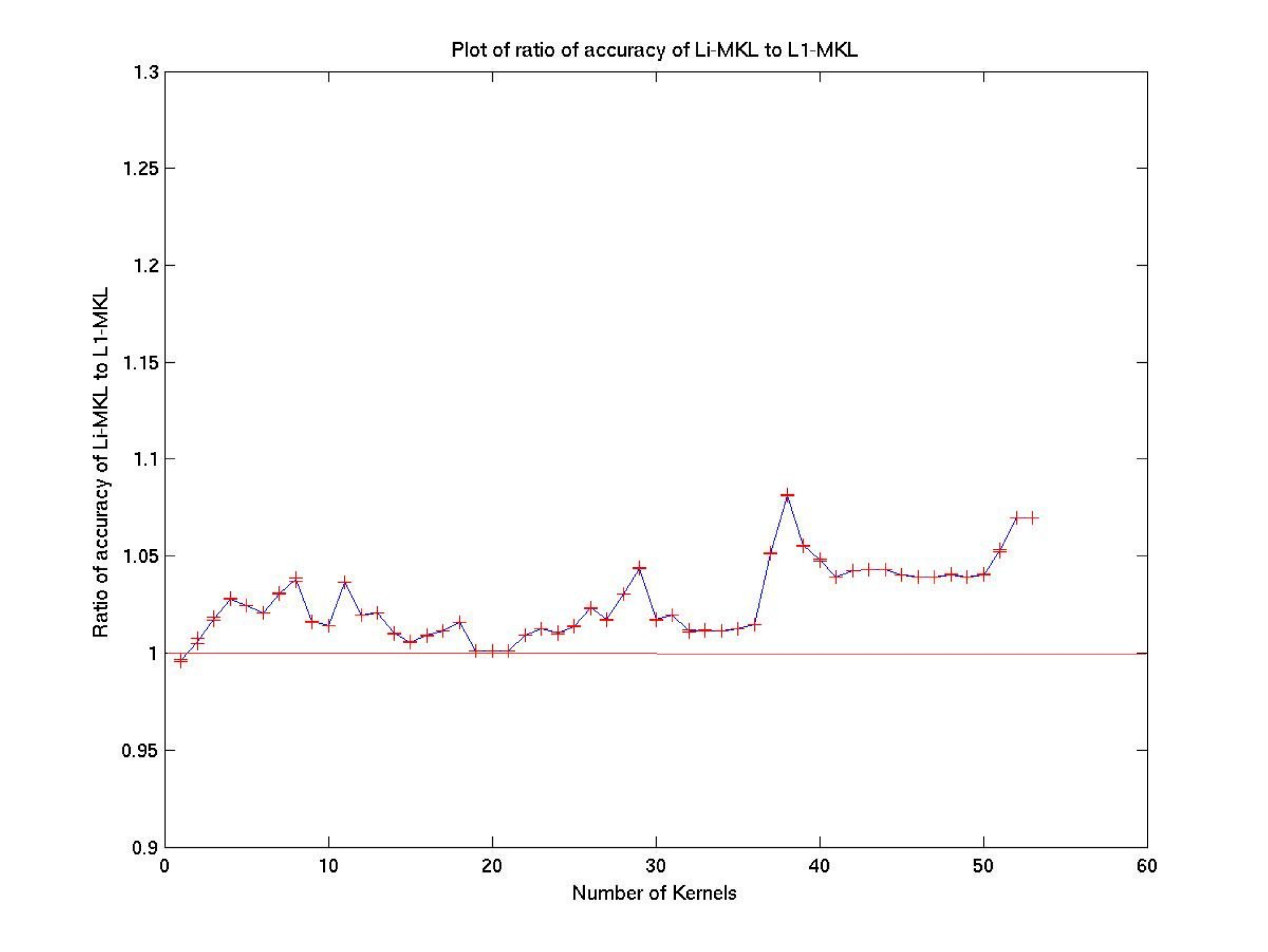}
\caption{Figure shows ratio of overall accuracies of \textbf{Li-MKL} to \textbf{L1-MKL} as function of number of kernels on Oxford flowers datasets.}\label{fig:ratio}
\end{figure}

\begin{figure}
\centering
\includegraphics[width=2in,height=2in]{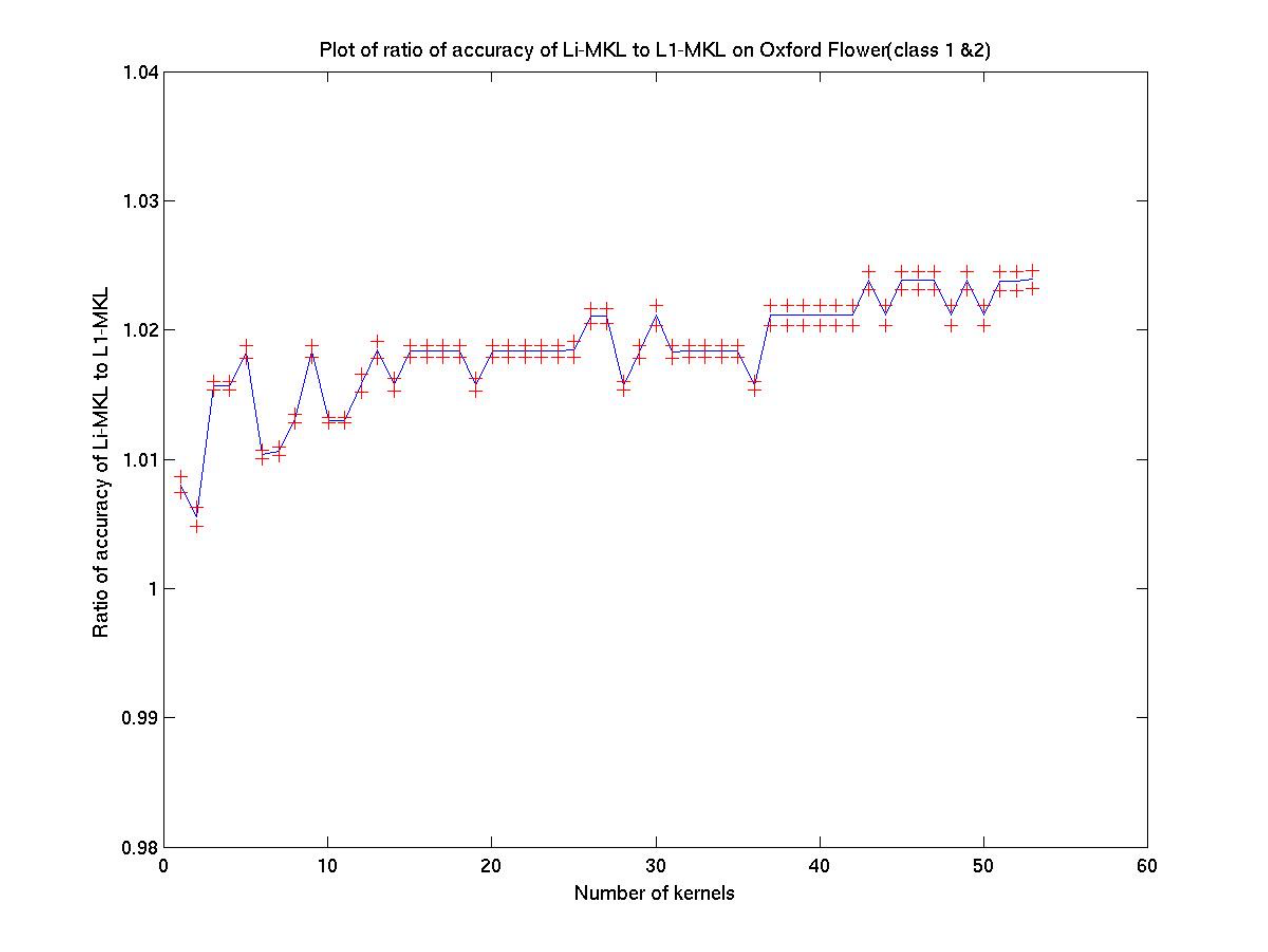}
\caption{Figure shows ratio of overall accuracies of \textbf{Li-MKL} to \textbf{L1-MKL} as function of number of kernels on some binary classification on Oxford dataset(Faces and Bikes).} \label{fig:ox_1_2}
\end{figure}
\begin{figure}
\centering
\includegraphics[width=2in,height=2in]{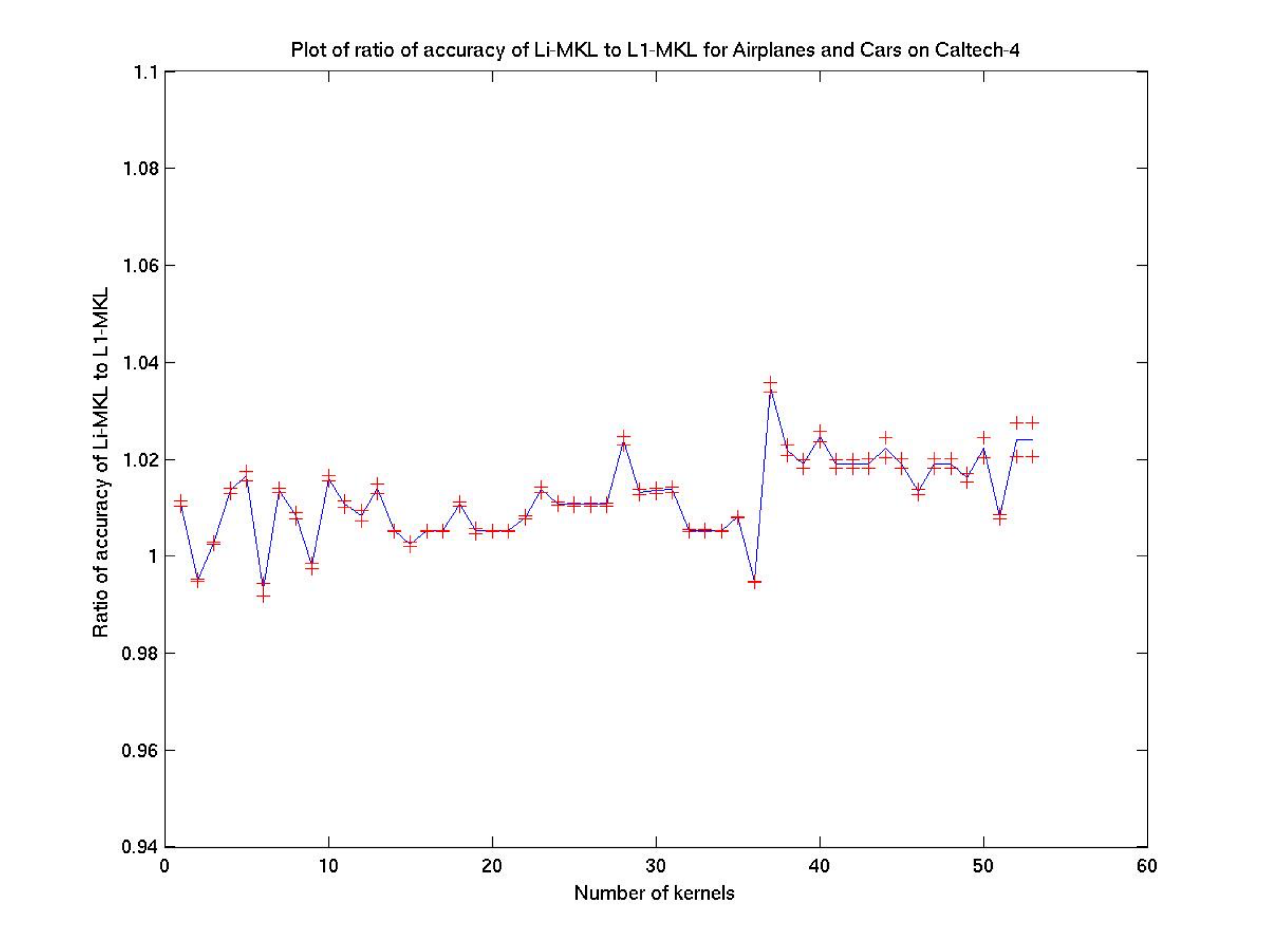}
\caption{Figure shows ratio of overall accuracies of \textbf{Li-MKL} to \textbf{L1-MKL} as function of number of kernels on some binary classification on Oxford dataset(Faces and Bikes).} \label{fig:ox_2_5}
\end{figure}

In the subsequent set of experiments the generalization performance of the \textbf{Li-MKL} and \textbf{L1-MKL} as a function of the number of base kernels is compared. The plots are shown in figures~\ref{fig:ratio} ~\ref{fig:ratio4} for the two benchmark datasets. Figures show that in most of the cases \textbf{Li-MKL} achieves better generalization than \textbf{L1-MKL}. Also, in some cases the improvement is as high as 7.5\%. Note that the base kernels were derived from the fixed sets of descriptors and hence have some degree of redundancy. These results show that the proposed formulation does achieve good improvement generalization even in these cases.
The next set of experiments compare the performance of the methodologies at various values of the regularization parameter $C$ (see figure~\ref{fig:c_ox}). Note that at performance of \textbf{L1-MKL} drastically decreases for low values of C. In some cases the difference in accuracy between \textbf{Li-MKL} and \textbf{L1-MKL} is as high as 9\%. Hence, the proposed formulation is less sensitive to the variation in the regularization parameter.

\subsubsection{Results on Caltech-101 dataset}
This section presents results on Caltech-101\footnote{\url{http://www.vision.caltech.edu/Image_Datasets/Caltech101/}}. Caltech-101 dataset contains 101 object categories. We have taken 30 images for each class, of which 15 are randomly taken as the training/validation data and the remaining as test data. We have used Pyramid Histogram Of Gradient (PHOG) features generated\footnote{Code available at \url{http://www.robots.ox.ac.uk/~vgg/research/caltech/phog.html}} at various levels (1,2,3) and angles (180,360). We have generated kernels on these six PHOG features using different parameters for the polynomial and Gaussian kernel (9 for each feature, totally 54 kernels). This experimental procedure was repeated for 3 times with different training-test data splits. The mean testset accuracies obtained with \textbf{L1-MKL} and \textbf{Li-MKL} were 31.45\% and 27.12\% respectively. This shows that the \textbf{Li-MKL} achieves better generalization. 

\subsection{Results of CKL Experiments}
All the experiments in this section is carried out using descriptors available from ColorDescriptor software\footnote{\url{http://staff.science.uva.nl/~ksande/research/colordescriptors/}}. The general procedure for the experiment is given in the figrue~\ref{fig:color}. All the experiments in this section follows:
\begin{figure*}
\centering
\includegraphics[width=5in,height=4in]{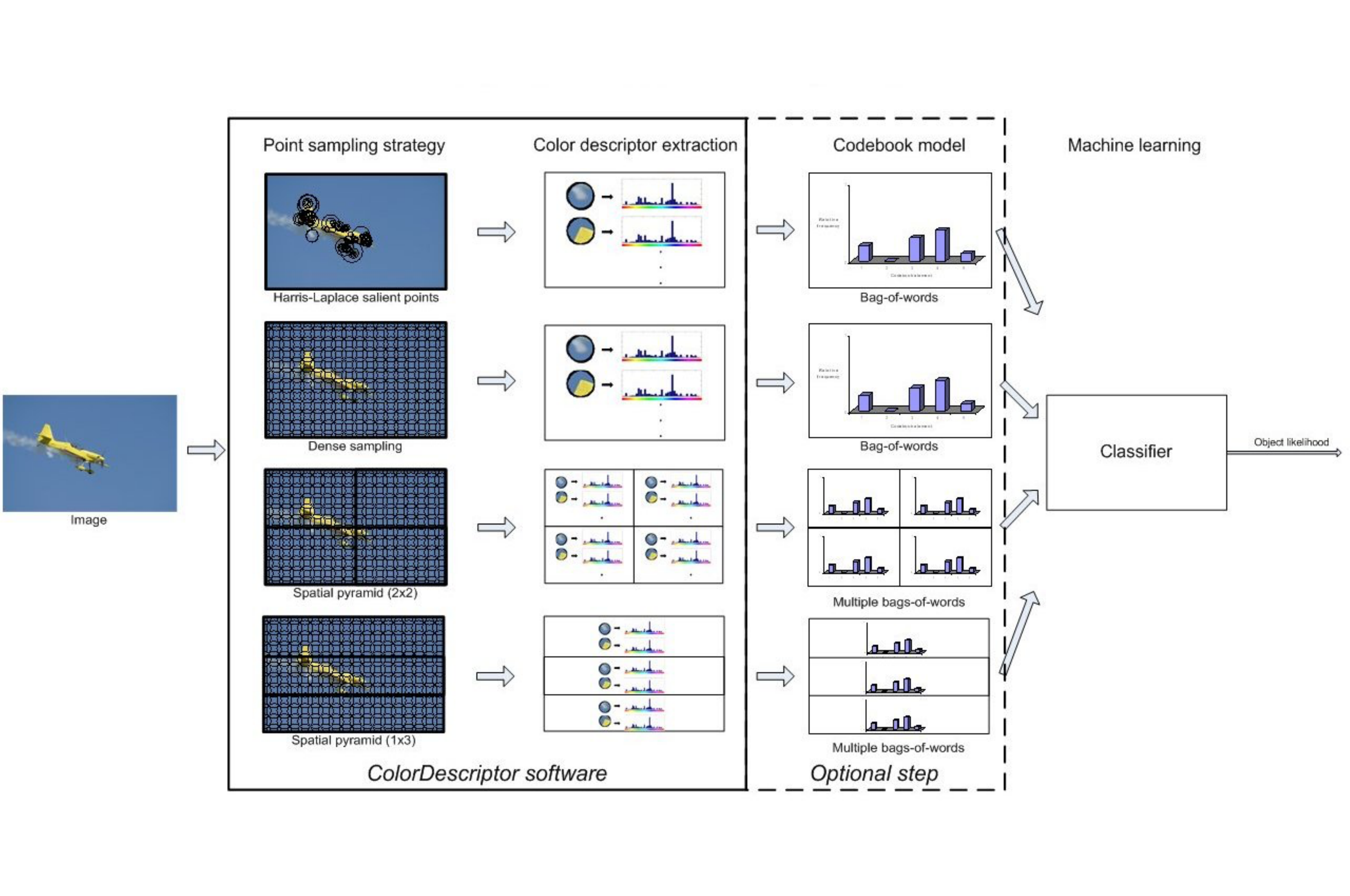}
\label{fig:color}
\end{figure*}

\begin{itemize}
 \item Generate  for all training images.
 \item Generate all the 14 descriptors provided by the software(RGB histogram, Opponent histogram, Hue histogram, rg histogram,  Transformed Color histogram, Color moments, Color moment invariants, SIFT, HueSIFT, HSV-SIFT, OpponentSIFT, rgSIFT, C-SIFT,  Transformed Color SIFT).
 \item Use no spatial pyramids.
 \item Clusters all points from all training images to form a codebook for each descriptors.
 \item Generate histogram of codebook of both training and testing images.
 \item Train the classifier using these histograms as features.
\end{itemize}
Results obtained using above procedure is given in following sections.
\subsection{Results on Caltech-5 dataset}
\begin{figure}
\centering
\includegraphics[width=3in,height=2.5in]{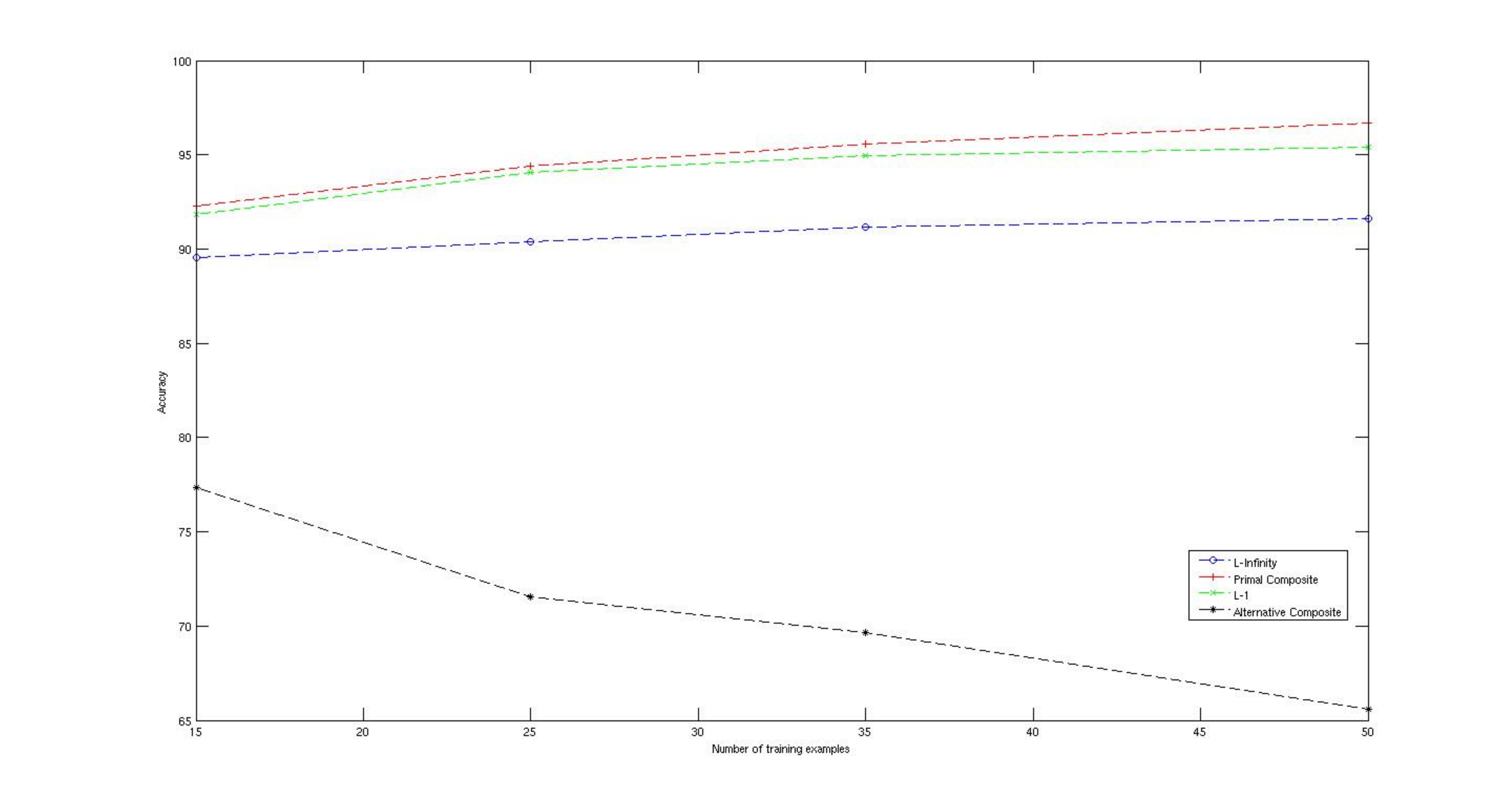}
\label{fig:mixed}
\caption{Shows plot of mean accuracy as number of training size.}
\end{figure}

This section presents results on Caltech-5\footnote{\url{http://www.robots.ox.ac.uk/~vgg/data/data-cats.html}} using new MKL formulation and descriptors(csift, opponentsift, rgsift, sift, transformedcolorsift) provided from ColorDescriptor software. Caltech-5 dataset contains images of airplanes, cars, faces, leopards and bikes. This section follows same procedure discussed in previous section. We have used cluster size of 100 to form codebook. Clusters are found using k-means algorithm. Note here we have not run k-means multiple times to find the best cluster or codebook. We have taken 100 images for each class, of which 15 are randomly taken to form a codebook. We have generated kernels on 5 descriptors provided using different parameters for Gaussian kernel (10 for each descriptor, totally 50 kernels). This experimental procedure was repeated for 5 times with different training-test data splits. Figure ~\ref{fig:mixed} reports mean accuracy as number of training size increases. Note here we have used same codebook generated on 15 training points for all training sizes. \\

\subsection{Results on Oxford dataset}
The task in the Oxford flower dataset is to categorize images of 17 varieties of flowers. This dataset contains 80 examples for each class. In~\cite{nils08}, the authors introduced four different features color, SIFT for foreground region, SIFT for foreground boundary, Histogram of Gradients for flowers. We have used the $\chi^2$ distances given in~\cite{nils08,nils06}\footnote{\url{http://www.robots.ox.ac.uk/~vgg/data/flowers/17/index.html}} for our experimentation on this dataset. We have used same training, validation and test splits as used in ~\cite{nils08}. The mean testset accuracy achieved by \textbf{L1-MKL}, \textbf{Li-MKL} and CKL are 85.3922\%, 86.6667\% and 86.6667\% respectively. 

\subsection{Results on Caltech-101 dataset}
This section presents results on Caltech-101\footnote{\url{http://www.vision.caltech.edu/Image_Datasets/Caltech101/}} using new MKL formulations and all 14 descriptors provided from ColorDescriptor software. We have taken 30 images for each class, of which 15 are randomly taken as the training/validation data and the remaining as test data. We have generated kernels on 14 descriptors provided using different parameters for Gaussian kernel (2 for each descriptor, totally 28 kernels). With cluster size 600 accuracy obtained is around 24.1\% and with cluster size 300, accuracy obtained is 23.21\%. Main problem in Caltech-101, is problem of clustering for forming codebook. Because of huge size and dimension in data, we followed 2-level kmeans. Our guess is that codebook formed is not good because clustering is not good.

\section{Conclusions and Future Work}\label{sec:conc}
The project addressed the issue of combining various descriptors for a given object categorization problem in order to achieve better generalization. The project also briefly addressed problem of video change detection. 

Adaboost has been designed for combining descriptors. State-of-the-art methodologies for object categorization employ a $l$-1 regularization based MKL formulation, which is more suitable for selecting descriptors rather than combining them. The key idea is to employ a $l$-$\infty$ regularization and mixed $l$-$\infty$ and $l$-1 regularization for combining the descriptors in an MKL framework. The new MKL formulation is better suited for object categorization and highly efficient algorithms which solve the corresponding convex optimization problem were derived.

Empirical results performed on synthetic and real-world benchmark datasets clearly establish the efficacy of the proposed MKL formulation. In some cases, the increase in accuracy when compared to the standard $l$-1 regularization was as high as $9\%$. The results also show that there is a consistant improvement in accuracy in almost all the cases, however, the improvement is maximized when the redundancy in the base kernels is low. Another advantage with the proposed formulation is that it is less sensitive to variation in the regularization parameter, $C$.

Work is going on to experiment the new MKL formulations for Caltech-101 dataset using codebook models described. Experiments is going on to form codebooks with different size as size of codebook largely affect classification accuracy. Novality of new MKL formulations will be known once experimentation on the bigger dataset has been done namely Pascal and Caltech-256. Future work also includes to experiment the new MKL formulations for these bigger datasets.

{\small
\bibliographystyle{ieee}
\bibliography{egbib}
}

\end{document}